\documentclass[
twocolumn,
]{ceurart}

\usepackage{listings}

\usepackage{enumitem}
\setlist[itemize]{leftmargin= 5.5mm}

\begin{document}

\copyrightyear{2023}
\copyrightclause{Copyright for this paper by its authors.
  Use permitted under Creative Commons License Attribution 4.0
  International (CC BY 4.0).}

\conference{Proceedings of the Sixth Workshop on Automated Semantic Analysis of Information in Legal Text (ASAIL 2023), June 23, 2023, Braga, Portugal.
}

\title{Enhancing Pre-Trained Language Models with Sentence Position Embeddings for Rhetorical Roles Recognition in Legal Opinions}

%
\author[1]{Anas Belfathi}[%
email= anas.belfathi@edu.univ-paris13.fr,
]
\address[1]{Nantes Université, École Centrale Nantes, CNRS, LS2N, UMR 6004, F-44000 Nantes, France}

\author[1]{Nicolas Hernandez}[%
email=nicolas.hernandez@univ-nantes.fr,
]
\cormark[1]

\author[1]{Laura Monceaux}[%
email=laura.monceaux@univ-nantes.fr,
]

\cortext[1]{Corresponding author.}

\begin{abstract}
The legal domain is a vast and complex field that involves a considerable amount of text analysis, including laws, legal arguments, and legal opinions. Legal practitioners must analyze these texts to understand legal cases, research legal precedents, and prepare legal documents. The size of legal opinions continues to grow, making it increasingly challenging to develop a model that can accurately predict the rhetorical roles of legal opinions given their complexity and diversity. In this research paper, we propose a novel model architecture for automatically predicting rhetorical roles using pre-trained language models (PLMs) enhanced with knowledge of sentence position information within a document. Based on an annotated corpus from the LegalEval@SemEval2023 competition, we demonstrate that our approach requires fewer parameters, resulting in lower computational costs when compared to complex architectures employing a hierarchical model in a global-context, yet it achieves great performance. Moreover, we show that adding more attention to a hierarchical model based only on BERT in the local-context, along with incorporating sentence position information, enhances the results.
\end{abstract}


\maketitle

\section{Introduction}


Pre-trained language models, such as BERT \cite{devlin2018bert} and GPT3 \cite{Bert2020GPT3}, have shown significant improvements in performance across various Natural Language Processing (NLP) tasks. However, when it comes to apply these models to specific domains like legal documents, unique challenges arise. Legal documents are often lengthy and without explicit structure, requiring the identification of 
coherent parts, known as Rhetorical Roles (RRs) for tasks such as summarization, information extraction, and legal reasoning  \cite{saravanan2008automatic,Bhattacharya2019RRs,malik2021semantic}.

In this research work, we are interested in the task of rhetorical role prediction in legal judgements. In that context, examples of RRs are: 
PREAMBLE (meta-data related to the legal judgment document), FACTS
(chronology of events that led to filing the case), RPC (final decisions ruled by the present court), etc. 
In particular, we work with the dataset provided by the organizers of the SemEval 2023 LegalEval competition for the rhetorical role prediction task\footnote{ \url{https://sites.google.com/view/legaleval}.}. 
For this task, Hierarchical Sequential Labelling Network (HSLN) \cite{jin2018hierarchical,Bhattacharya2019RRs,brack2021sequential,kalamkar2022corpus} and Pre-trained Language Models (PLM) which are able to handle long word sequence \cite{beltagy2020longformer,zaheer2020bigbird,ouyang2021ernie}  are prefered. 
However, simple text statistics on the data show that a text contains on average 4346.07 sub-words\footnote{Computed with the BERT tokenizer.} ($\pm$2151.08)  and therefore exceeds the maximum input length any Pre-trained Language Models can handle. 
In addition, the best current system does not exceed 87\% of F1-score which justifies an interest in the task \cite{modi2023legaleval}.

Our main contributions are the following:
\begin{itemize}
\item we enhance the pre-trained language model BERT with sentence position information at input;
\item we study the sentence position information under various representations (absolute, normalized and K–quantile);
\item we consider two architectures to contextualize the sentence representations: 1) a single BERT encoder and 2) a hierarchical model made of a BERT encoder layer to encode sentences and a shallow encoder (with two Transformer layers) to contextualize a sentence with its surrounding sentences; 
\item we evaluate these various models in the context of a rhetorical role sequence labelling task for legal judgments.
\end{itemize}

Our related work section (Section~\ref{secrelatedword}) covers various topics from the fusion of discourse information in language models to some rhetorical role prediction system architectures including a position embedding presentation. Then, in Section~\ref{secproposedmodels}, we detail the models we propose. In Section~\ref{sectmethodology}, we present the methodology we use for our fine-tuning evaluation. Finally, we present our result and discuss our future work in the last sections.

Our code will be publicly available on MASKED\_URL.

\section{Related work}
\label{secrelatedword}

\textbf{Injecting discourse information in language models}
Legal texts share linguistic characteristics specific to the legal domain (and often to a legal sub-domain). They have legal jargon, long sentences, unusual word order and long length \cite{zhong2020benefit,chalkidis2020legalbert,kalamkar2022corpus,chalkidis22lexglue}. 
These characteristics do not allow to take full advantage of the state-of-the-art language models trained on the general domain and even show their limitations since most of them cannot handle a text length which goes beyond their maximum input length.

Transformers \cite{vaswani2017attention} suffer from a quadratic computational and memory complexity with respect to the sequence length. This lead most of the SOTA models (e.g. BERT \cite{devlin2018bert}, RoBERTa \cite{liu2019roberta}, LegalBERT \cite{chalkidis2020legalbert}) to adopt 
 512 
 as their maximum sequence length.

Pre-training or retraining with discourse-based objectives can result in sentence and text representation that are  more adapted to addressing NLP tasks at the discourse level \cite{jernite2017discoursebasedobjectives,iter2020pretrainingContrastiveSentenceObjectives,yang2020smith} 
but this kind of approach does not address directly the limitation of the input length.

In terms of neural architecture adaptation, Hierarchical Attention Networks (HANs) have been proposed to model a sequence of sentences by stacking two layers of encoders: one to capture the word sequences and another (taking the former as input) to capture the sentence sequence. This architecture has been shown to perform significantly better than single layer encoders for text classification \cite{yang2016hierarchical}, text segmentation \cite{lukasik2020textSegmentation}, recommendation \cite{yang2020smith} and sequence labelling tasks \cite{kalamkar2022corpus}. 
To avoid an important number of padded words in the first layer, \cite{yang2020smith} proposed to concatenate as many as natural sentences the input block can fit. 
Although extending the scope of single transformer encoder, the HAN architecture does not solve the complexity problems and has to process fixed length of sentence sequence and truncate too long documents. 
Recent architectures such as Longformer \cite{beltagy2020longformer}, BigBird \cite{zaheer2020bigbird} or Ernie \cite{ouyang2021ernie} succeeded in extending to 4096 the maximum sequence length while reducing the Transformer's complexity\footnote{See \cite{fournier2023lighter} for a survey of techniques to address Transformer's limitations.} by introducing sparsity into attention layers (i.e. by allowing each token position to attend to a subset of token positions with respect to some sparse patterns). 
To improve such models, \cite{chalkidis22lexglue} suggested to consider the actual logical structure of documents.

Apart from these more complex architectures, \cite{lukasik2020textSegmentation} showed that, for a  text segmentation task, looking at the local context around each candidate break (by taking the end of the previous sentences and the beginning of the following as input) is sufficient to obtain comparable performance to the HAN architectures.

\textbf{Rhetorical Role Prediction} 
The analysis of the rhetorical text\footnote{We do not make a distinction here between the intra- and inter-sentencial levels.} structure includes several tasks \cite{lippi2016argumentation,yamada2019legalArgumentationInJapanese,putra2021parsing,kalamkar2022corpus,modi2023legaleval}: 1) text segmentation into text rhetorical units, 2)  rhetorical role identification of each text unit, 3) structure prediction, which links the text units together and 4)  relations labelling to name the connections. 
We focus here in the task of rhetorical role identification which has been considered in the literature as a  
sequence labelling task taking the sentence as a minimal text unit \cite{modi2023legaleval}.

As reported by \cite{modi2023legaleval}, the Hierarchical Sequential Labelling Network (HSLN) remains the most efficient architecture for this task (at least for the LegalEval dataset) \cite{jin2018hierarchical,Bhattacharya2019RRs,brack2021sequential,kalamkar2022corpus}. The model first encodes sentence representations (e.g. by using sent2vec \cite{pagliardini2018unsupervised} or any PLM like BERT or LegalBERT). Then it contextualizes the sentence representations (e.g. through a BiLSTM layer) and eventually predicts the label sequence thanks to a sequence labelling layer (like a CRF).
On the LegalEval dataset, best performance were obtained by participants who used domain-adaptation techniques like a pre-train language model trained on Legal text, or augmented datasets. The baseline model were based on the HSLN architecture and had a performance of 79\% F1 score. The proposed methods
show an improvement over the baseline without ever succeeding to outperform by more than seven points.
\cite{cohan2019pretrained} showed that a single BERT can be sufficient to capture contextual dependencies without the
need for hierarchical contextual encoding neither a CRF sequence labeler. 
The approach uses BERT to encode a concatenation of sentences (fixed at 10 sentences) and use a MLP over each encoded sentence separator token to predict the corresponding sentence label. 
Despite its low complexity, the approach is limited by the length of the input sequence and requires to tile the whole text to obtain the whole label sequence. \cite{kalamkar2022corpus}  did not confirm the effectiveness of the method on the LegalEval dataset.

\textbf{Position embeddings}
By nature, 
the Transformer architecture is 
not sensitive to the order of input tokens. To make the model position-aware, the position information of the input words is typically added as an additional embedding to the input token embeddings \cite{vaswani2017attention}. 
While absolute sinusoidal position encodings were  utilized in the vanilla Transformer, 
some works showed that learned position embeddings can provide more flexibility in adapting to different tasks through back-propagation \cite{devlin2018bert, gehring2017convolutional}, 
instead of using hand-crafted position representations \cite{wang21positionEmbeddings, lin2022survey}. 
Multiple works explored also different token position information (absolute, relative) and ways to include it in Transformers (e.g. in the input or the attention matrix) \cite{wang21positionEmbeddings,chen21positionalEcoding}. 
Very few were interested in sentence position information. \cite{yang2020smith} indicate to fuse sentence block representations with sentence block position embeddings but without mentioning precisely the nature of the position (relative, absolute...) and how the position table is built.

\textbf{BERT Input Representation} 
BERT is one of the models that utilizes three types of learned embedding layers (See Figure~\ref{fig:mesh1}): Token Embeddings, Segment Embeddings, and Position Embeddings \cite{devlin2018bert}:

\begin{figure}
    \centering
\includegraphics[width=\textwidth/2]{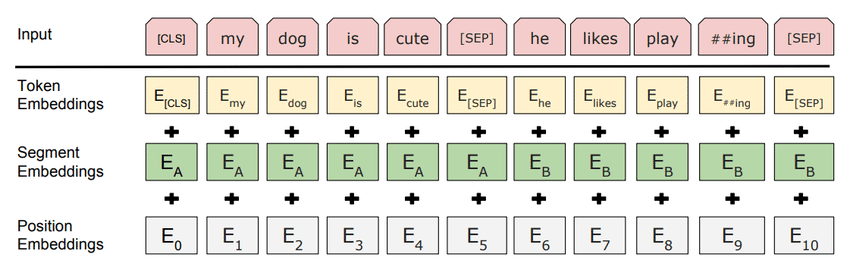}
    \caption{BERT input representation. The input embeddings are the sum of the token embeddings, the segmentation embeddings and the position embeddings\cite{devlin2018bert}.} 
    \label{fig:mesh1}
\end{figure}

\begin{itemize}
\item \textbf{Token Embeddings:} This layer is responsible for converting each word in the input text into a fixed-dimensional vector representation. In the BERT Base model, each word is represented as a 768-dimensional vector.
\item \textbf{Segment Embeddings:} This layer has the task of distinguishing between the inputs in a given pair by assigning one of two vector representations to each token in the input.
\item \textbf{Position Embeddings:} BERT takes into account the sequential nature of input sequences by learning a vector representation for each token position in the input. The Position Embeddings layer is a lookup table of size (512, 768), where each row represents the vector representation of a word at a specific position in the input sequence.
\end{itemize}

The representations from Token Embeddings, Segment Embeddings, and Position Embeddings are summed element-wise to produce a single representation with shape (1, n, 768), where $n$ is the length of the input sequence. This combined representation captures contextual information of tokens in the input text. Although BERT has shown effective results for many tasks such as question answering and sentiment analysis of tweets, it is non-performant when working with lengthy documents. However, it does not incorporate any information about the position of sentences within a document, which can be crucial for identifying the RR of a particular sentence. To address this limitation, we propose the addition of sentence position embeddings to the BERT embedding, which aims to enhance the performance of RR prediction in legal opinions.

\section{Fusion sentence position embeddings at input}
\label{secproposedmodels}

In this section, we focus on the approach that we have developed for injecting discursive knowledge without pre-training through input embeddings. The use of input embeddings has been a popular approach in natural language processing (NLP) for representing text data in a high-dimensional vector space. By incorporating discursive knowledge into these embeddings, we aim to improve the performance of NLP models without the need for pre-training.

\subsection{BERT-SentPos: BERT sentence encoder enhanced with sentence position embeddings at input}
\label{sectionBERTSentPos}

\begin{figure}
  \centering
  \includegraphics[scale=0.3]{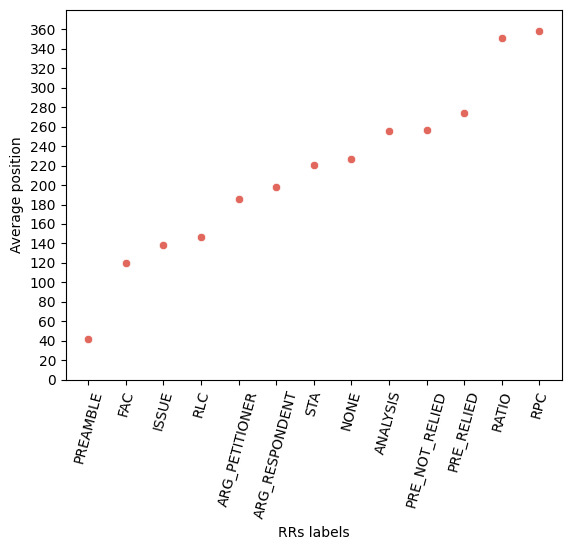}
  \caption{Average sentence position variation by label on the dataset.}
  \label{fig:sp}
\end{figure}

In the analysis we have conducted on legal documents, we have observed that each rhetorical role has a specific position within the document (See Figure~\ref{fig:sp}). For example, the preamble role is found at the beginning, the analysis role in the middle of the document, and the RPC role at the end. To improve our model's performance, we decided to incorporate additional information that indicates the position of each sentence in the document. We achieved this by adding an extra embedding layer to the BERT embedding, which helped us capture the sequential nature of positions in vector space. Various Position Embeddings (PEs) have been proposed in Transformer based architectures \cite{vaswani2017attention} to "capture the sequential nature of positions in vector space." These PEs range from fixed ad-hoc ways to fully learnable ones. BERT, in particular, uses fully learnable PEs. In the interest of simplicity, we decided to reuse the learned PEs to represent the sentence Position Embeddings. By doing so, we were able to create a more accurate and effective model for analyzing legal documents.

In this research, we employed various techniques to examine the positioning of sentences within legal documents. Specifically, we explored three different ways of analyzing sentence positions, including absolute position, normalized position, and k-quantile position :

\textbf{Absolute position} refers to the location of a sentence within a particular document in relation to other sentences in the same document.  \textit{For instance}, we may have a document that includes:
\begin{itemize}[noitemsep, topsep=0pt, partopsep=0pt, parsep=0pt, label=\labelitemiv]
    \item Sentence 1: "The court hereby orders the defendant to appear for a hearing on Monday."
    \item Sentence 2: "The defendant shall provide all relevant documents to the plaintiff's attorney by Friday."
\end{itemize}

In this case, Sentence 1 has an absolute position of being the first sentence in the document, while Sentence 2 has an absolute position of being the second sentence.

\textbf{Normalized position} refers to the process of converting the position of a sentence in one document to a corresponding position in the largest document in a corpus that has the maximum length of sentences. This is done to ensure consistent normalization across different documents, by aligning sentence positions with respect to a common reference point. 
\textit{For example}, let's consider Document A, which consists of 50 sentences, and Document B, the largest document in the corpus, containing 100 sentences. If we take Sentence 25 from Document A, we can calculate its normalized position as (25 multiplied by 100 divided by 50) = 50. This aligns Sentence 25 with the same relative position in Document B.

\textbf{K--quantile position} The analysis of legal documents often involves addressing variations in document length and unique writing styles of judges. This technique involves dividing the documents into k parts or quantiles to help control the absolute position of each rhetorical role. \textit{For example}: We have an Document X (divided into 4 quantiles):

\begin{itemize}[noitemsep, topsep=0pt, partopsep=0pt, parsep=0pt, label=\labelitemiv]
    \item Quantile 1: "The court presents the background of the case."
    \item Quantile 2: "The court discusses relevant legal precedents."
    \item Quantile 3: "The court analyzes the evidence presented by both parties."
    \item Quantile 4: "The court renders its final decision and issues the judgment."
\end{itemize}

By dividing the opinion into quantiles, each containing a specific role or aspect, it helps in addressing variations in document length and unique writing styles of judges.

\begin{figure}
    \centering
    
\includegraphics[scale=0.55]{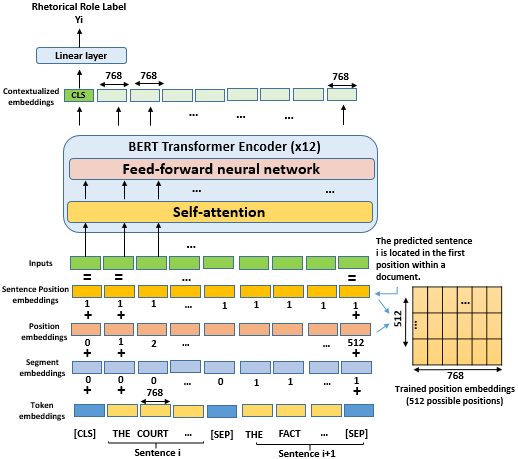}

    \caption{BERT-SentPos: A BERT Architecture with Sentence Position Embeddings fused at the BERT input layer using the Learnable Position Embedding of Tokens.}
    \label{fig:mesh2}
\end{figure}

Figure~\ref{fig:mesh2} illustrates our proposed architecture, which utilizes the BERT encoder. We input pairs of consecutive sentences denoted as sentence $i$ and sentence $i+1$, along with the corresponding sentence position of predicted sentence $i$, to enable our model to learn contextualized representations that consider the relationships between neighboring sentences. To capture positional information, we added BERT embeddings with sentence position embeddings. These embeddings provide the model with position information of each sentence within the document. The combined representation is passed through encoders and feed-forward layers for predicting the rhetorical role of sentence $i$. At the end we obtained three models: BERT enhanced with Absolute Sentence Position Embeddings (BERT-AbsPos), BERT enhanced with 8-Quantile Sentence Position (BERT-8-QuantilePos), BERT enhanced with Normalized Sentence Position (BERT-NorPos).

\subsection{HiBERT: Hierarchical variant to contextualize sentence representations}

\begin{figure}
  \centering
  \includegraphics[scale=0.4]{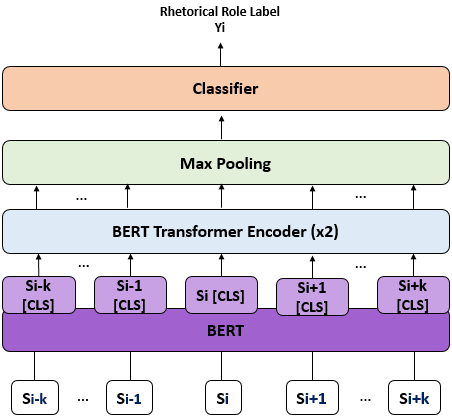}
  \caption{HiBERT: A shallow hierarchical encoder over a BERT-SentPos encoder to capture the sentence sequences coherence. 
  The input layer takes the same input as the BERT-SentPos namely, for each sentence, an aggregation of token, segment, position and sentence position embeddings.}
  \label{fig:HierBERT}
\end{figure}

In this section, we present HiBERT, a hierarchical variant of BERT-SentPos described in Section~\ref{sectionBERTSentPos} (See Figure~\ref{fig:HierBERT}). The model is based on the \cite{chalkidis2020legalbert,chalkidis-etal-2021-paragraph}'s hierarchical model.
The model aims to label a sentence according to the sentences that precede and follow at a certain range. 
Each sentence representation is enhanced with sentence positional information before being encoded by a BERT encoder. 
The process generates a top-level representation, denoted as $S_i[CLS]$, for each sentence. 
These sentence representations are then fed into a  2-layered transformer encoder to contextualize a sentence sequence. 
This encoder follows the same specifications as BERT, including hidden units and the number of attention heads.
Eventually we utilize a dense layer to predict the label of the sentence in focus in the current sentence sequence.

We have set a window of surrounding sentences to a maximum of $\pm$7 sentences. This range size corresponds to the average number of consecutive sentences with the same RR label in our train dataset.
Four variations of HiBERT were experimented: HiBERT-AbsPos, which uses Absolute Sentence Positions with the maximum window size; HiBERT-NorPos, which uses Normalized Sentence Position with the maximum window size; HiBERT-AbsPosHalf, which uses Absolute Sentence Position with half of the maximum window size; and HiBERT-NorPosHalf, which uses Normalized Sentence Positions with half of the maximum window size.  
Overall, the hierarchical model with more attention attention is an effective solution for processing legal documents that consist of thousands of words. By using a hierarchical approach and taking into account the context of the text, the model is able to effectively process lengthy documents and make accurate classifications based on the document's content.



\section{Experimental methodology}
\label{sectmethodology}

We evaluate our contributions in the fine-tuning phase of pre-trained models.

\begin{table}[t]
    \caption{Corpus Statistics: The corpus is split into training set, validation set and test set. The table shows number of documents, sentences and the maximum sentence position}
  \label{tab:cor-stat}
  \begin{tabular}{cccc}
    \toprule
    \textbf{Dataset} & \textbf{Docs} & \textbf{\begin{tabular}[c]{@{}c@{}}Sentences\end{tabular}} & \textbf{\begin{tabular}[c]{@{}c@{}}Max Sentence\\ Position\end{tabular}} \\
    \midrule
    Train            & 220           & 25866                                                          & 385                                                                      \\
    Validation       & 25            & 2875                                                           & 348                                                                      \\
    Test             & 30            & 2849                                                           & 208                                                                      \\
    \midrule
    \textbf{Total}   & 275           & 31590                                                          &                                                                          \\
    \bottomrule
  \end{tabular}

\end{table}

\subsection{The LegalEval Dataset}

We utilized the data supplied by Sub-task~A ``Rhetorical Roles Prediction'' of the SemEval~2023 Task~6 ``LegalEval - Understanding Legal Texts'' challenge\footnote{ \url{https://sites.google.com/view/legaleval}.}. 
The dataset comprises Indian legal data extracted from court judgments and includes 13 different RRs, with the details and definitions for each RR provided in the article by Kalamkar et al. \cite{kalamkar2022corpus}. 
The average number of sentences per document is 117.31. 

To prepare the dataset, we kept the same LegalEval split (train and validation data). 
We used the 90\% of the train data to train and the remaining 10\% to validate the model. 
Furthermore, we used the original validation data to evaluate the performance of the trained models. The statistics for splitting the corpus are shown in Table~\ref{tab:cor-stat}. 

\subsection{Hyperparameters}

For the fine-tuning setup, hyperparameters are determined through experimentation and analysis. The batch size is set to 8, taking into consideration the available computational resources and model performance. The learning rate is set to 2e-5, which is a commonly used value for fine-tuning NLP models. The epoch number is chosen from \{1, 2, 3\}, with the final epoch number selected based on a balance between training time and model performance for each model.

\subsection{Performance measures}

The performance of the NLP models for the rhetorical roles task is assessed using Weighted-Precision ($_wP$),  Weighted-Recall ($_wR$), Accuracy ($A$), Weighted-F1 ($_w{F1}$) and Macro F1 ($_M{F1}$) scores based on the hidden test set. The weighted F1 score considers both precision and recall, and it is calculated by taking into account the class-wise F1 scores weighted by the number of samples in each class. The Macro F1 score provides an overall assessment of model performance by calculating the F1 score for each class independently and then taking the average.

\section{Results}

\begin{table}[t]
  \caption{Performance of Models on Test Data}
  \label{tab:mod-perf}
  \centering
  \setlength\tabcolsep{4pt} 
  \begin{tabular}{lccccc} 
    \toprule
    \textbf{Model} & \textbf{$_wP$} & \textbf{$_wR$} & \textbf{$A$} & \textbf{$_w{F1}$} & \textbf{$_M{F1}$} \\
    \midrule
    BERT-HSLN & {\bf 0.79} & {\bf 0.81} & {\bf 0.81} & {\bf 0.79} & 0.57 \\
    \midrule
    BERT & 0.69 & 0.67 & 0.67 & 0.65 & 0.47 \\
    BERT-8-QuantilePos & 0.73 & 0.74 & 0.74 & 0.72 & 0.52 \\
    BERT-AbsPos & 0.75 & 0.74 & 0.74 & 0.74 & 0.55 \\
    BERT-NorPos & 0.76 & 0.77 & 0.77 & 0.75 & {\bf 0.58} \\
    \midrule
    HiBERT-AbsPos & 0.76 & 0.74 & 0.74 & 0.75 & 0.56 \\
    HiBERT-NorPos & 0.76 & 0.77 & 0.77 & 0.75 & 0.54 \\
    HiBERT-NorPosHalf & 0.77 & 0.76 & 0.76 & 0.75 & 0.53 \\
    HiBERT-AbsPosHalf & 0.77 & 0.79 & 0.79 & 0.77 & 0.54 \\
    \bottomrule
  \end{tabular}
\end{table}

In this section, we present the experimental results of our study on injecting different types of sentence position embeddings in the BERT model (See Table~\ref{tab:mod-perf}). As mentioned earlier, we used the weighted F1 score as our primary performance metric.  
As a baseline, we report the score obtained by the BERT-HSLN\footnote{This work is part of OpenNyAI  \url{https://opennyai.org/} mission, which is funded by the EkStep Foundation \url{https://ekstep.org/}.} model \cite{modi2023legaleval} 
which achieved a performance of 79\%. 
Our first experiment was conducted using BERT with three types of sentence position embeddings: Absolute position (BERT-AbsPos), Normalized position (BERT-NorPos), and K-quantile position(BERT-8-QuantilePos). Our results revealed that BERT performs comparatively poorly (65\%) compared to the proposed models when considering three types of position. For the K-quantile position, we experimented with different numbers of parts and found that the best division based on performance is with 8 parts. However, we found that BERT with normalized position achieved a better score of 75\%. We attribute this improvement to our efforts in controlling the variation in length across different legal documents. 

To further enhance the performance of the recognition of RRs, we experimented with a hierarchical model that combines BERT with more attention and a window size equal to the average of consecutive sentences with the same RR label, while also injecting sentence position information (HiBERT-AbsPos and HiBERT-NorPos). Unfortunately, we did not observe any significant improvement of the results. 
Subsequently, we halved the average window size of sentences to take into consideration (HiBERT-AbsPosHalf and HiBERT-NorPosHalf). This led to an improvement in the results, particularly with Absolute position by 79\%. 
We attribute this success to the fact that absolute position captures the specific position of a sentence within a document, providing crucial contextual information of the predicted sentence based on its surrounding sentences. This improvement with a smaller context window can be explained by the fact that the semantics of a sentence are usually more dependent on local context than on knowing all sentences in a paragraph, for example. 

Overall, our experiments demonstrated that using different types of sentence position information can significantly improve the performance of BERT on legal document classification tasks. Additionally, a hierarchical model that combines BERT with absolute position information and a window size of 4 sentences ($\lceil 7/2\rceil$) can further enhance the performance of our proposed models. Furthermore, our approach also achieved low computational time, making it efficient and practical for real-world applications.

\section{Conclusion and Future Work}

The results of our study indicate that the inclusion of absolute, normalized, and k-quantile positional embeddings can significantly improve the performance of both BERT base and its hierarchical variant for the Rhetorical Roles Prediction Task. 
However, there is a room for improvement. 
It should be noted, for example, that one potential limitation of this approach is that the number of sentences in a given document may exceed the dimensions of the learnable embedding matrix. The framework proposed by \cite{wang21positionEmbeddings} can help us to determine the interest of learning dedicated sentence position embeddings or using sinusoidal PEs.
In addition, in the context of token position embeddings, \cite{chen21positionalEcoding} showed that encoding position to attention matrix per-head results in superior performance comparing to adding position embeddings to the input. A similar experiment could be conducted on the sentence position embeddings. 
Looking ahead, our future work aims to develop a new architecture with a greater number of layers that incorporates additional features such as metadata pertaining to each legal opinion. Furthermore, we plan to explore the potential benefits of pre-training instead of fine-tuning and fine-tuning with LegalBERT \cite{chalkidis2020legalbert}, a BERT-based model pre-trained on legal text. Our motivation for this direction is based on the fact that current state-of-the-art models may not be sufficient to achieve optimal performance for legal NLP tasks. By incorporating these improvements, we hope to develop a more robust architecture that achieves optimal performance for the Rhetorical Roles Prediction Task and other legal NLP tasks.

\begin{acknowledgments}
This research was funded, in whole or in part, by l’Agence Nationale de la Recherche (ANR), project ANR-22-CE38-0004. 
\end{acknowledgments}

\bibliography{legal}

\end{document}